%% file: main.tex
\documentclass{article}

\usepackage{booktabs} %
\usepackage{numprint}
\usepackage{xcolor}
\usepackage{microtype}
\usepackage{hyperref}
\usepackage{amsfonts}
\usepackage{amssymb}
\usepackage{enumitem}

\usepackage{graphicx}
\usepackage{subfigure}
\usepackage{amsthm}
\usepackage{amsmath}

\newcommand{\footremember}[2]{%
    \footnote{#2}
    \newcounter{#1}
    \setcounter{#1}{\value{footnote}}%
}
\newcommand{\footrecall}[1]{%
    \footnotemark[\value{#1}]%
} 

\newtheoremstyle{myplain}{2pt}{2pt}{}{0pt}{}{.}{5pt plus 1pt minus 1pt}{}
 
\newtheoremstyle{sty2}%
{1pt}%
{1pt}%
{}%
{}%
{}%
{}%
{}%
{}%

\newtheoremstyle{slanted}
  {1pt}%
  {1pt}%
  {\slshape}%
  {}%
  {\bfseries}%
  {.}%
  {0.5em}%
  {}%

\theoremstyle{slanted}

\theoremstyle{definition}
\newtheorem{definition}{Definition}[section]
 
\theoremstyle{remark}

\theoremstyle{slanted}
\newtheorem*{example}{Example}

\npthousandsep{,}\npthousandthpartsep{}\npdecimalsign{.}

\title{Explaining an increase in predicted risk for clinical alerts}

\author{%
  Michaela Hardt\footremember{g}{Google}%
  \and Alvin Rajkomar\footrecall{g} \footremember{ucsf}{UCSF}%
  \and Gerardo Flores\footrecall{g}%
  \and Andrew Dai\footrecall{g}%
  \and Michael Howell\footrecall{g}%
  \and Greg Corrado\footrecall{g}%
  \and Claire Cui\footrecall{g}%
  \and Moritz Hardt\footnote{UC Berkeley, work done at Google}%
  }

\begin{document}
\maketitle

\begin{abstract}
Much work aims to explain a model's prediction on a static input. We consider explanations in a temporal setting where a stateful dynamical model produces a sequence of risk estimates given an input at each time step. When the estimated risk increases, the goal of the explanation is to attribute the increase to a few relevant inputs from the past. 

While our formal setup and techniques are general, we carry out an in-depth case study in a clinical setting. The goal here is to alert a clinician when a patient's risk of deterioration rises. The clinician then has to decide whether to intervene and adjust the treatment. Given a potentially long sequence of new events since she last saw the patient, a concise explanation helps her to quickly triage the alert.

We develop methods to lift static attribution techniques to the dynamical setting, where we identify and address challenges specific to dynamics. We then experimentally assess the utility of different explanations of clinical alerts through expert evaluation.
\end{abstract}

\input{intro}
\input{problem_statement}

\input{usecase}

\input{related_work}

\input{approach}

\input{linear}
\input{experiments}

\input{conclusions}

\section{Acknowledgements}
We would like to thank our colleagues 
Emily Xue, Been Kim, Jake Marcus, Chloe Hsu, 	
I-Ching Lee, Jimbo Wilson, James Wexler, Hector Yee, Yi Zhang.

\bibliographystyle{unsrt}
\bibliography{references}
\newpage
\input{appendix}

\end{document}

%% file: intro.tex
\section{Introduction}

Routinely framed as a static prediction task, statistical risk assessment is often a dynamic problem. Risk estimates evolve over time as new measurements arrive and additional data become available. Alerts may be triggered once the risk or the increase in risk has reached a critical threshold. An alert can depend on numerous events in the past, making it difficult to quickly understand which events contributed to the increase in risk. The goal of our work is to provide tools to quickly assess an increase in predicted risk in dynamical risk assessment scenarios.

Our main application is a clinical early warning system that alerts physicians to the deteriorating health of a hospitalized patient~\cite{Bates2015-tv,Bates2014-zj}. Such a system can decrease mortality and length-of-stay~\cite{RRintervention}. It combines events such as measured vital signs, laboratory tests, and notes from doctors into a patient risk score. Once the risk score exceeds a threshold, an alert notifies the doctor. Alerts should have high precision -- too many false alerts can lead to fatigue~\cite{Drew2014-ge}. Alerts should also be informative;  a clinician must be able to quickly assess what factors are contributing to the patient's increase in risk to identify interventions. This is not an easy task as there may have been dozens or hundreds of new events since the clinician last saw the patient. Simply reporting abnormal lab results or vitals is unhelpful in situations where abnormalities are common. For example, in the case of critically ill patients, abnormal lab results may be unrelated to the alert and are often nonspecific~\cite{Fernando2018-ae}.

While much work on model interpretability has focused on identifying important features of a static input, these works do not address salient temporal effects, such as how the relevance of input changes over time as newer data become available. Our work begins to address these important questions with a detailed use case in the medical domain.

\textbf{Our contributions.}
We broadly study explanations of an increase in risk in a dynamic setting. This is a departure from the predominantly studied static setting and we hope that it will spark more interest in the future.
Within this broader context, we dive deep into an application and use our methods to explain alerts in a clinical early warning system, where we see our results as providing a set of valuable baselines for future work. More specifically, we make the following contributions:

\begin{itemize}[noitemsep,topsep=0pt,parsep=0pt,partopsep=0pt]
 \item Identify an important, yet neglected, question: How can we explain \emph{changes} in prediction over time? 
 \item Develop methods to lift commonly used static gradient-based attribution techniques~\cite{GradientImage,visualization_techreport,Gradients,Sundararajan2017AxiomaticAF} to the dynamical risk assessment setting. 
 \item Analyze our methods in the simplified theoretical model of a linear dynamical system with a quadratic risk function to form an analytic understanding of the semantics and challenges.
 \item Implement our methods in the context of an early warning system to explain alerts, and open-source our code, see supplementary material. 
 \item Evaluate our methods through expert judgment by medical students and an ICU doctor who were asked to assess the clinical utility and compare them to attention~\cite{Bahdanau2014-iu} and statistical methods~\cite{Caruana:2015:Pneumonia,Rothman2013-zx}. 
\end{itemize}

\textbf{Organization.}
We describe the problem setup and its application to a clinical early warning system in Sec.~\ref{sec:problem_statement}. Related work is reviewed in Sec.~\ref{sec:related_work}.
We then present our methods in Sec.~\ref{sec:method-der} and analyze them theoretically in Sec.~\ref{sec:linear} in a simplified linear dynamical system gaining insights into the challenges that motivate some refinements.
Sec.~\ref{sec:experiments} details our experiments. We conclude in Sec.~\ref{sec:conclusions}.

%% file: problem_statement.tex
\section{Problem Statement}
\label{sec:problem_statement}
We now introduce the general problem we address. 
At each time step $t$ with $t=0,1,2,\dots,T$, a real-valued vector, $x_t\in\mathbb{R}^d$, is provided as input to a stateful model to produce an updated risk estimate $p_t\in\mathbb{R}$. We denote by $h_t$ the hidden state of the model at time $t$. 
We then phrase the problem of explaining an increase in predicted risk as the identification of a few values in the sequence of inputs that are relevant to a risk increase. Clearly, the whole sequence of inputs perfectly explain the prediction, however, there may be too many inputs for a human to make sense of. Also some inputs may rather be indicative of a decrease in risk and would not make good explanations. The notion of relevancy is domain specific and depends on the consumers~\cite{RigorousInterpretability}. We therefore assess it through human evaluation in our experiments. 

\textbf{Notation.} In this paper we consider attribution weights, $a\in\mathbb{R}^{d\times T}$, over all inputs and the explanations are then simply the input values with the highest weights.
The weights may vary depending on which risk increase from $p_{t_0}$ to $p_{t_1}$ we are trying to explain.  This can be made explicit with a superscript $a^{t_0 \rightarrow t_1}$.  For instance, we may expect weights from the future $t > t_1$ to be zero.

\subsection{Use case of an early warning system}\label{sec:usecase}

%% file: usecase.tex
\begin{figure}
\begin{subfigure} %
  \centering
  \includegraphics[width=0.55\textwidth]{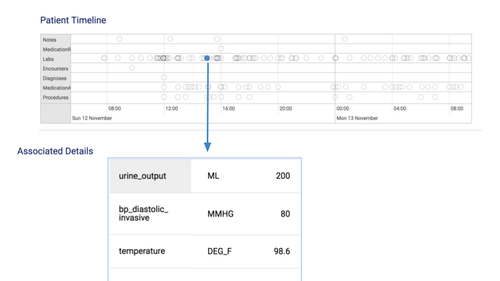}
\end{subfigure}
\begin{subfigure} %
  \centering
  \includegraphics[width=0.45\textwidth]{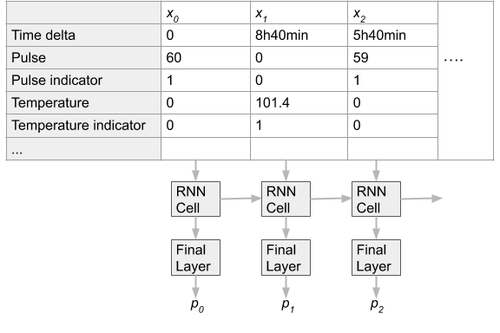}
\end{subfigure}
  \caption{Data from a patient in a health system ordered by time. On the left we see the timeline of diagnoses, procedures,  medications, etc. of a patient with the lab results at a particular time expanded. On the right we focus on the representation of irregular lab results and show how they can be used as input into a recurrent neural network depicted in unrolled format.}
\label{fig:pipeline}
\end{figure}
One promise of digital health is that electronic health record (EHR) data of hospitalized patients could be continuously monitored by electronic systems that alert physicians when a patient's health is worsening~\cite{Bates2014-zj,Bates2015-tv}. Early warning systems for deterioration have already been shown to decrease mortality and length-of-stay~\cite{RRintervention}. Since many patients are hospitalized because they are at high risk, repeated alerts for persistently elevated risk may induce alarm fatigue, a known clinical problem. Instead, alerts are likely more relevant when a patient's risk has increased, especially over a short period of time, where there may exist possible interventions to blunt or reverse the increase in risk~\cite{Rothman2013-zx}.
For example, we may seek to trigger alerts for patients whose mortality risk increased by at least 50\% and is now above 0.2.
In practice, a hospital can tune these alerts based on  precision (or the number needed to evaluate) and recall (or the available resources). 

Given an alert, a clinician has to decide quickly what to do about it in the presence of many competing inputs and responsibilities~\cite{Block2013-ko,Mamykina2016-zn}. A concise explanation of an alert is helpful for triaging of alerts and  clinical decision making. A good explanation should focus on the new information since the clinician last saw her patient.

\textbf{Input Data.} A patient's timeline from heterogeneous EHR data including diagnoses, medications, procedures, notes from doctors, vitals and lab results is illustrated in 
Fig.~\ref{fig:pipeline}. 
While all of these features may be relevant for the risk assessment, they do not all make good explanations. Medications, for example, may be associated with a high risk without causing it. For instance, Norepinephrine is administered to patients in shock and therefore accompanies but actually decreases their high risk of dying.  
As a first step, we focus on lab results and vitals as explanations.
For concreteness, we describe a commonly used input representation~\cite{ChePCSL16:MissingvalsRNN,Lipton2016DirectlyMM,Suresh2017-yr}, see Fig.~\ref{fig:pipeline}. We set $x_{t,i}$ to be the normalized result of a lab test  $i$ at time $t$ (or 0 when the lab test was not taken). To distinguish a lab result of zero from an absent result we further include an indicator for each lab  test. We also track the time between two consecutive events as a feature.
The reason to focus on lab results is twofold. First, there is an abundance of lab results compared to any other type of medical event in the EHR, see Fig.~\ref{fig:pipeline}, which makes it easy to miss important ones. They can be collected routinely but may only be reviewed by a clinician during the morning rounds or evening sign-outs.  Second, they can give us insights into the physiological state of the patient, with a caveat discussed next.

\textbf{Challenge of confounded measurements.} The existence of a lab result is often due to a worried clinician ordering  a test~\cite{Agniel2018-em}. Patterns of lab tests can be as revealing as their results~\cite{testmeasures,Lipton2016DirectlyMM} and increasing the frequency of measurements leads to a risk increase~\cite{Suistomaa2000}.
Furthermore, results are affected by treatments as the following example illustrates.

\begin{example}
The tidal (lung) volume of 500ml of a patient was associated with a high risk despite the fact that 500ml is normal. This is partially because this test is only performed on intubated patients (generally at higher risk). But even after comparing only intubated patients the value of 500ml was still associated with an increased risk.  The ventilation settings of a patient can be either specified by a target volume (commonly 500ml) or by pressure (in which case case the volume varies more widely). The first method is chosen more often for serious cases and once patients get better they may be switched to the second method. Hence, a value of 500ml is associated with more serious cases requiring intubations.
\end{example}
While our methods are general and do not depend on this specific input and domain, we will discuss their ability to deal with confounded irregular measurements.

%% file: related_work.tex
\section{Related Work}
\label{sec:related_work}
The increasing use of complex models for decision making in critical areas including health care and the criminal justice system, has raised the question of model interpretability: We need to be able to check the soundness of the reasoning of our models~\cite{RigorousInterpretability}. For a general overview of interpretability see the comprehensive survey~\cite{InterpretabilitySurvey}.

\textbf{Explaining Predictions.}
Gradient-based methods have been applied  mostly in image classification to produce a saliency map~\cite{GradientImage,visualization_techreport,Gradients,smoothgrad,Deconv,Sundararajan2017AxiomaticAF,GuidedBackprop,QAIntGrad}. Attention-mechanisms have been used with recurrent neural networks (RNNs)~\cite{Bahdanau2014-iu}. 
Techniques requiring only black-box access to a model use local additive models~\cite{Ribeiro:2016:WIT:2939672.2939778}, ablate features through Shapley values~\cite{Shapley}, or maximize the mutual information of the predictions and the explanations~\cite{expl_mutual_inf}. 

\textbf{Model Interpretability for Health Care.}
In clinical practice, mostly small and simple models based on 1-5 features are deployed~\cite{Mandell2007-lc, Lim2015-ps,Churpek2016-aq,Howell2012-ih}.
While easy to inspect, their accuracy is limited.
With rich data now digitized in EHRs~\cite{Adler-Milstein2015-ak}, more features and model architectures are being developed with improved accuracy~\cite{Shickel2017-ef}. These complex models raise  questions of interpretability: How can we understand these models~\cite{pmlr-v68-suresh17a,Nguyen2016-hj, Choi2016-qo,Che:deep-phenotyping,Lasko}? How can we group  patients~\cite{Nguyen2016-hj,Tran:2015:representation, Doshi-Veleze54}? How can we explain predictions? 
To explain predictions on EHR data (the focus of this paper), a variety of approaches have been explored starting with generalized additive models~\cite{Caruana:2015:Pneumonia} over discretized features, or feature crosses fitted using gradient boosted decision trees. Subsequent work has extracted more complex discretized features that also incorporate temporal aspects using the maximum information gain criteria~\cite{interpretability-alerts-cardio}.
To explain a patient's risk, statistics of these discretized features can be used  such as the odds ratio or the Rothman index~\cite{Rothman2013-zx}.
Other lines of work have studied latent Dirichlet allocation~\cite{TopicModels}, 
convolutional neural networks with feature ablation~\cite{Avati2017-rg}, RNNs with an attention mechanism~\cite{Sha:distillation,Retain-Attention}, and co-distillation~\cite{Che:deep,Che:distillation}.

Our paper differs from previous work by explaining an increase in predicted risk rather than explaining a static risk prediction. 
Our methods build up on prior work and lift them to the dynamical setting.

%% file: approach.tex
\section{Proposed Dynamical Gradient-Based Methods}
\label{sec:method-der}
In this section we describe how we can lift several existing static attributions to the dynamical setting. Some adaptions are rather simplistic, while others provide a new perspective on the dynamics. 
Our methods rely on a model that is trained to predict the risk $p_t$ given inputs $x_0,\dots,x_{t}$ and take various derivatives of the predictions $p_t$.

\paragraph{Input Gradients.}
A natural method~\cite{GradientImage,visualization_techreport,Gradients} to explain the risk $p_{t_1}$ considers the derivative of the risk score with
respect to an input~$x_t$ at a time step~$t\le t_1:$
\[
a_t^{t_1} = \frac{\partial p_{t_1}}{\partial x_t}.
\]
This captures how a small change in the input affects the prediction.
To explain a change in predictions we can apply a time restriction to the time window of interest.
The underlying assumption is that good explanations of a change in risk from $p_{t_0}$ to $p_{t_1}$ should contain recent events in $(t_0, t_1]$.
\begin{definition}\label{def:time-restr}
Given a sequence of attribution weights $a_0, a_1, \dots$, we define the \emph{time-restricted explanations} of the change in prediction between $t_0$ and $t_1$, $a^{t_0\rightarrow t_1}_t$, at time $t$ as $a_{t}$ for $t_0 < t \le t_1$ and $0$ otherwise.
 \end{definition}
There are several popular variants
that improve on the basic input gradient method~\cite{smoothgrad,Deconv,GuidedBackprop,Sundararajan2017AxiomaticAF}. In the following, we will devise variants with a greater focus on the dynamical aspect of the problem.

\paragraph{Temporal Integrated Gradients.} 
We extend the \emph{integrated gradient} method~\cite{Sundararajan2017AxiomaticAF,QAIntGrad} that averages out the gradients along the line
segment between two chosen input sequences, the target sequence $\hat{x}=(\hat{x}_0,\dots)$ and a baseline sequence
$b=(b_0,\dots)$, typically set to all zeros.
Formally, the path-integrated gradient of a prediction, $p_{t_1}$, is given by the integral
\[
S(b, \hat{x})=(\hat{x}-b)\int_0^1 \left.\frac{\partial p_{t_1}}{\partial
x}\right|_{x=\alpha b+(1-\alpha)\hat{x}}\mathrm{d}\alpha\,.
\]
We lift path-integrated gradients to the dynamical setting by purposefully constructing a suitable baseline. This allows us to explore how intermediate values between the old and the new ones would have affected the prediction as motivated by the following example.

\begin{example} If a patient currently has a temperature of 102F changing it a bit may not change the risk very much. However, if we compare it with the value the patient had the last time the doctor came around, say 98F, and compute the gradient of interpolations, the predicted risk may be sensitive to values close to the fever threshold of 100.4F.
\end{example}

Specifically, we construct a baseline $b=(b_0, \dots, b_{t_1})$ given the actual input sequence $\hat{x}=(\hat{x}_0, \dots, \hat{x}_{t_1})$ as follows. For early time steps $t \leq t_0$, we set $b_t = \hat{x}_t$. 
Afterwards, for $t > t_0$, the baseline pretends that the results have not changed since $t_0$, that is we substitute recently measured values of a feature after $t_0$ with the latest measurement of that feature until time $t_0$.
To address the challenge of irregular measurement patterns, we consider an input $\hat{x}$ in which not all features are measured each time. We construct $b$ by keeping the exact same measurement patterns of $\hat{x}$, replacing all results after $t_0$ with the most recent measurement up until $t_0$.

In our example, for temperature measurements of 99F at noon, 100.1F at 1pm, 100.2F at 2pm, 100.9F at 3pm, the baseline to explain the change in risk between 1:30 and 3pm is $99, 100.1, 100.1, 100.1$ copying forward the last value before 1:30pm. With other intermittent measurements the sequence could look like  $99, \bot,\bot, 100.1, \bot, 100.2, 100.9$ with a baseline of  $99, \bot, \bot, 100.1, \bot, 100.1, 100.1$.

\begin{figure}
\begin{subfigure} %
  \centering
  \includegraphics[width=0.5\textwidth]{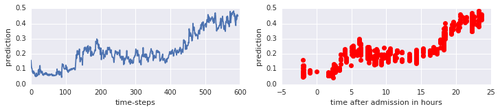}
\end{subfigure}
\begin{subfigure} %
  \centering
\includegraphics[width=0.5\textwidth]{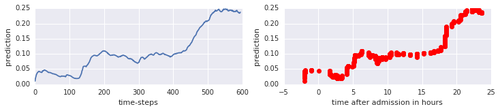}
\end{subfigure}
\vskip -0.15in
  \caption{A patient's risk time series predicted from RNN models on the left trained without a smoothing loss term and on the right trained with smoothing.
  The first and third graph use time steps as x-axis, while second and fourth use the time relative to admission illustrating clustered activity. 
  }
\label{fig:smoothing}
\end{figure}

\paragraph{Time Derivatives.}
Another natural approach considers the derivative of the risk score with respect to time, $\frac{\partial p_t}{\partial t}$. 
The naive discrete time approximation of this derivative is
$p_t-p_{t-1}$. This allows us to assign a weight to each time-step. We setup our input so that there is exactly one feature $i_t$, $0\leq i_t < d$, present in the input at step $x_{t}$. Then we can set $a_t = (p_t-p_{t-1}) \cdot e_{i_t}$ with absent features receiving a $0$ weight. 
The motivation is similar to that of time restrictions: A good explanation of a change in predictions from $p_{t_0}$ to $p_{t_1}$ should contain events between $t_0$ and $t_1$. At the limit, this means that the change $p_{t} - p_{t - 1}$ should be attributed to the event at time $t$.

\textbf{Properties:}  The weights of events are consistent in time. That is even as future events come in, the weights of past events do not change.
Moreover, the sum of the weights in $(t_0,t_1]$, $\sum_{t_{0} < t \leq t_1} a_t$ is equal to $p_{t_1} - p_{t_0}$, as it is a telescoping sum.

\textbf{Challenge:} The discrete time derivatives perform poorly in the presence of noise and large sampling intervals, as is well known and has motivated much work in signal processing, see, e.g., \cite{signal-diff}.

To reduce noise, we suggest to alter the training objective of the model. Instead of focusing solely on minimizing the loss of the prediction $p_t$ compared to the actual risk, we additional minimize the changes of the prediction over time by adding a smoothing loss of the form~$\eta\sum_t (p_t-p_{t-1})^2$ for some scalar~$\eta>0$. We refer to this refined method as \emph{smoothed discrete-time derivatives}. The loss term promotes a smoother sequence reducing noise. 
Fig.~\ref{fig:smoothing} illustrates shows the same patient's predicted risk time series with and without smoothing. With a smoothing loss term and $\eta=0.005$ much of the noise is removed while the shape is retained.
Notably, the absolute risk is lower with smoothing which may not affect the explanations since they rely on the differences in risk. %

%% file: linear.tex
\section{Analysis in a linear dynamical system}
\label{sec:linear}

Although our methods apply to general differentiable time-series models, we ground our discussion here in a simple illustrative example of a linear dynamical system. Recall at each time step $t$, an
input~$x_t\in\mathbb{R}^d$ influences the evolution of the hidden
state $h_t$ of a stateful model and yields a new risk score $p_t$. In our linear dynamical system the risk score is
defined recursively as $p_t = 0.5 \|h_t\|^2$ and $h_t = Ah_{t-1} + Bx_t$.
In words, the risk score at time $t$ is the squared Euclidean norm of
a hidden state~$h_t\in\mathbb{R}^n$ that evolves from a known hidden
state $h_0$ according to a linear dynamical system specified by two
linear transformations~$A\in\mathbb{R}^{n\times n},
B\in\mathbb{R}^{n\times d}$. 
For the purpose of our example, we could replace $\|h_t\|^2$
with any quadratic form $h_t^\top Qh_t.$

\textbf{Time derivatives.} 
In a continuous version of our linear dynamical system with $\frac{\partial h_t}{\partial t} = Ah_{t-1} + Bx_t$ the continuous time derivatives are (by definition and the chain rule)
\[
\frac{\partial p_t}{\partial t} = h_t^\top A h_{t-1} + h_t^\top Bx_t
\]
It is very common to approximate continuous time derivatives through discrete approximations such as $p_t-p_{t-1}.$ However, such approximation tend to be highly noisy as has motivated much work on discrete differentiation schemes in the signal processing community, see~\cite{farid2004differentiation}.

\textbf{Input gradients.} 
Applying the multivariate chain rule reveals that in our linear system
\[
\frac{\partial p_{t_1}}{\partial x_t}= h_{t_1}^\top A^{t_1-t} B 
\]

The index~$t$ of the input only influences the power of the matrix~$A$. For fixed~$t$, the gradient can grow or shrink exponentially as $t_1$ increases. These phenomena known as \emph{exploding/vanishing gradients} lead to the gradient method emphasizing inputs far in the past or very recently. 
The issues of vanishing and exploding gradients can be mitigated through careful design of the model, e.g. the use of LSTMs~\cite{Hochreiter1997-gq}, and by the time-restriction to recent inputs.

\textbf{Integrated gradients.} We analyze the integrated gradient in our linear dynamical system. Denoting by $h_{t_1}[x]$ the hidden state that the system attains on an
input sequence $x=(x_1,\dots,x_{t_1}),$ the above expression simplifies using the linearity of the integral operator and the system's update rule to 
\[
S_t(b,x)=\left((h_{t_1}[b]+h_{t_1}[x])/2\right)^\top A^{t_1-t} B
\]
What we see is that the path-integrated gradient simplifies to the
arithmetic mean of the baseline gradient and the target gradient.
Although not too different from the basic gradient method in this
example, path-integrated gradients with carefully constructed baseline can alleviate some of the shortcomings of the basic gradients.

%% file: experiments.tex
\section{Evaluation}
\label{sec:experiments}
\subsection{Dataset}

We use data from the Medical Information Mart for Intensive Care (MIMIC-III)~\cite{mimic3} restricted to patients hospitalized for greater than 24 hours. MIMIC-III is widely accessible to researchers under a data use agreement. The data has been deidentified in accordance with Health Insurance Portability and Accountability Act (HIPAA) standards using structured data cleansing and date shifting.

We included more than \numprint{50000} hospitalizations, with patients randomly split into training (80\%), validation (10\%), and test (10\%) sets.   We consider the commonly studied prediction task of in-hospital mortality~\cite{Avati2017-rg,ChePCSL16:MissingvalsRNN}. We predict at various time points throughout the hospitalization how likely a patient is to die within this hospital stay. The in-hospital mortality rate was around 8\%. 
In Section~\ref{sec:exp-aki}, we study a different form of deterioration, namely acute-kidney injury (AKI).

\textbf{Input Data for Explanations.} 
For the models generating explanations we create a  whitelist of eligible lab tests (listed in the Appendix), because not all lab results are easily interpretable. For instance, the red blood cell distribution width is often picked up by predictive models, but not well understood \cite{FELKER200740}. Also among highly correlated lab tests, clinicians prefer one over the other (e.g. hemoglobin over hematocrit). We limit the lab values to the past three days. 

\subsection{Alert Generation}
We chose a model for alerting that is separate from the various models used to generate explanations. In particular, we use a sequence model similar to~\cite{Rajkomar2018-ae}.
That way we can decouple the quality of the alerts from the quality of the explanations. However, this means that our explanations may not be faithful to the alerts. %
We assume the clinician finished their rounds at 12 hours after admission at which point in time they have assessed the patient and developed a treatment plan. The risk predicted at this point is compared to the risk recomputed every 2 hours up until 24 hours after admission when the evening sign-out takes place and another team takes over.
We trigger an alert for a patient if their risk increases by at least a factor of 1.5 to at least a risk of 0.2. The cost of human evaluation limits us to one specific setup. 
We randomly selected patients that are being alerted on for the first time that have at least 40 new events. For patients with fewer new events a doctor could feasibly look over all of them.  We excluded subsequent alerts as the clinician may have snoozed the alert.

\begin{table*}[]
    \centering
    \begin{tabular}{l | l | l}
    & \textbf{Mortality Risk}  & \textbf{AKI Detection} \\
    Model & Precision @ 3  & Precision @ 1 \\
    \hline
    Random guessing & 0.24 [0.16, 0.33] & 0.1 \\
    Time-restricted gradients & 0.31 [0.23, 0.39] & 0.52  \\
    Time-restricted attention & 0.33  [0.23, 0.43]  & 0.3 \\
    Smoothed discrete-time derivatives & 0.40 [0.30, 0.50] & 0.47 \\
    Time-diffed odds ratio &  0.52 [0.39, 0.63] &  0.4\\
    Time-diffed Rothman index &  0.52 [0.42, 0.63] &  0.44 \\
    Time-restricted odds ratio &  0.52 [0.42, 0.62] &  0.53 \\
    \textbf{Temporal integrated gradients} &  \textbf{0.57}  [0.47, 0.68] &  \textbf{0.55} \\
    \end{tabular}
    \caption{For mortality risk, precision of the top-3 highest weighted lab results from distinct lab tests with a 95\% bootstrap confidence interval based on expert judgment. Precision of the highest weighted lab results for the detection of indicators of acute kidney injury.}
    \label{tab:expl}
\end{table*}

\subsection{Methods}
We compare our gradient-based explanation methods from Sec.~\ref{sec:method-der}, the standard gradients, the temporal integrated gradients with a carefully designed baseline, and the time derivatives, to a few methods from the literature.
An \emph{attention mechanism}~\cite{Bahdanau2014-iu} for recurrent neural networks derives weights for each input event to create a prediction. These weights can serve directly as explanations. We apply time-restriction in our dynamical setting.
We further include two statistics commonly used to associate risk with discretizied features: The \emph{odds-ratio}, e.g. in~\cite{Caruana:2015:Pneumonia},  and the \emph{Rothman index}~\cite{Rothman2013-zx}. To explain a change in prediction between time $t_0$ and $t_1$, we consider time-restriction (Def.~\ref{def:time-restr}) and \emph{time-diffing}. Time-diffing requires not just that the lab result falls in the time window $(t_0, t_1]$ of interest, but also that something about this test has changed since $t_0$. For example, a continuously high heart rate, although associated with a high risk, does not explain a change in risk. More formally, to compute the time-diffed weight of a lab result in $(t_0, t_1]$, we subtract the odds-ratio/Rothman index from results of the same test on or before $t_0$ from the current odds-ratio/Rothman index.  
We provide further details in the appendix. 
Additionally, a strawman, the \emph{random guessing} method randomly selects three recent inputs conditional on having three different features.

\paragraph{Gradient-based Methods.} 
We implemented an RNN using TensorFlow over normalized lab results (from which outliers were removed) with a single new lab result per step as illustrated in Fig.~\ref{fig:pipeline}.  
We take the time-series of observation values, together with an indicator of which lab test is present and an encoding of the time since the last step as input to the RNN as done previously in~\cite{ChePCSL16:MissingvalsRNN,Lipton2016DirectlyMM,Suresh2017-yr}. A notable difference is that no bagging is applied and for each step in the RNN only a single lab result is processed. We predict the label at each time step. Although this does not affect overall accuracy, it allows us to attribute changes in predictions to individual lab values for the time derivatives.
We use an LSTM~\cite{Hochreiter1997-gq} of size 64 with input (0.03), output (0.02) and recurrent (0.01) dropout~\cite{dropoutRNN}. For optimization we used the Adam optimizer~\cite{Adam} and a learning rate of 0.002 over batches of size 16 and clipped the gradients to 6. These hyperparameters were tuned with a proprietary Bayesian optimization framework. Training was performed using Tensorflow with Tesla P100 GPU. An open-source release of our code accompanies this manuscript.

\subsection{Metrics of the Expert Evaluation}
We obtain attribution weights for the techniques described above and select the top lab results from 3 distinct lab tests. Our raters included one ICU doctor and two medical students who jointly rated a total of 40 alerts labeling more than 800 lab results.
We gave them access to the patient's chart up until 12 hours after admission (around 5000 lab results). We asked to which extent the new lab results indicate that the patient's condition deteriorated since then. The new lab results that were selected as explanations (3 from each method) were presented in a random order to avoid position bias. We average the precision across patients. See the Appendix for details.

\subsection{Results for Mortality Risk Prediction}

\textbf{Quantitative Results of the Expert Evaluation.}
We compare the precision of the three highest weighted lab results from distinct lab tests in Table~\ref{tab:expl}. 
Overall we see that the temporal integrated gradients and the statistical explanations perform best with an average precision of the top-3 highest-weighted lab results above 0.5. 
The attention, gradients of the inputs, and smoothed discrete-time derivatives do not perform as well. Their average precision is better than random, however the confidence intervals overlap.

\input{case_study}

\subsection{Evaluation of AKI Explanations}\label{sec:exp-aki} 
To assess the robustness of our methods, we study a second task of AKI detection on the MIMIC-III dataset. 
The label is defined following  a subset of the KDIGO guidelines~\cite{kdigo}: If the serum creatinine increases by  at least 0.3 mg/dl within the past 48 hours or if the urine volume is less than 0.5 ml/kg/h (25ml/h, assuming 50kg weight) for 6 hours then we say the patient has a positive label. We train models making predictions every 3 hours during the hospital stay, excluding cases when a patient previously had a positive prediction within their encounter. 
Manual hyper-parameter tuning yielded best results for a learning rate of 0.0003 and a batch size of 64. Other hyper parameters remained the same and were not re-tuned. 
Note, this is a simple detection task and not a prediction of the future. Therefore, ground-truth explanations are available following the definition of the label. In particular, we consider an explanation correct when it selects any recent urine or creatinine value. 
Table~\ref{tab:expl} lists the top-1 precision for 100 randomly chosen positive examples. The temporal integrated gradients remain the best method overall. Other methods differ in effectiveness compared to the mortality prediction task. In particular, all gradient-based methods perform well.

%% file: case_study.tex
\textbf{A Qualitative Case Study.} 
\label{sec:case_study}
A qualitative account by an attending physician highlights strengths and weaknesses of the different methods in a case study of a particular patient.

\emph{In one patient presumed sepsis, the risk increased from 62.3\% to 86.2\%. The time-diffed odds-ratio selected a high temperature of 102.8 degrees Fahrenheit, low urine output, and an oxygen level of 100\%.  We suspect that such a high oxygen level is more commonly seen in invasively ventilated patient (e.g. with a breathing tube), so it indicates risk indirectly.  The human rater selected the first two as convincing evidence of a risk increase. However, we note that in each case, the prior values were all abnormal: the prior temperature was 100.8,  oxygen saturation was 100\%, and the urine output had been previously low.}
\emph{In the same case, the temporal integrated gradient technique selected low diastolic (55 mmHg) blood pressure, low mean (74 mmHg) blood pressure, and low urine output.  These blood pressures, in absolute terms, are somewhat low  but not below clinical thresholds that would necessarily require emergent action on their own.  However, in the context of this patient, they are relatively low, which does require urgent evaluation to determine the source of clinical change. The prior values were normal at at 72 and  95 mmHg, respectively and had been at those levels for multiple hours. Human raters selected all three values as convincing evidence.}

\emph{This case study highlights that the odds-ratio technique is excellent at selecting very abnormal values as evidence of increased risk.  If the odds-ratio, however, is not adjusted, then we did see likely confounding affecting the thresholds: a high oxygen level indicated high risk because it was confounded by ventilation across the entire dataset. The integrated technique seems, on qualitative inspection, to better select changes in vital signs and lab values with respect to the patient's personal baseline.}

%% file: conclusions.tex
\section{Conclusions and future work}
\label{sec:conclusions}
While much work on model interpretability has focused on explaining risk in a static setting, we introduced a new problem of explaining changes in predicted risk. Our new methods lift static gradient-based attribution techniques~\cite{GradientImage,visualization_techreport,Gradients,Sundararajan2017AxiomaticAF} to this dynamical setting. 

We applied our methods to explain clinical alerts of increased mortality risk by identifying three important recent lab values and compare them to  attention~\cite{Bahdanau2014-iu}, odds ratio~\cite{Caruana:2015:Pneumonia} and Rothman index~\cite{Rothman2013-zx}.
In our experiments we found that temporal integrated gradients and lifting statistical methods had the highest precision of above 0.5. Whether this will translate to useful explanations for clinical decision making has to be determined through a user study~\cite{RigorousInterpretability} in a clinical setting. Before a deployment, work is needed to ensure that the explanations are reliable~\cite{UnreliabilitySaliency} and neither misleading~\cite{Lipton16a:Mythos} nor creating unjustified trust~\cite{Springer:Trust}.
Our study is limited to providing explanations for an increase in risk. Understanding the interaction of model and explanation quality, e.g. whether explanations help identifying false alerts, is left for future work.
This work is a first step towards explaining changes in predicted risk and we hope it sparks further ideas, improvements and applications.

%% file: appendix.tex
\setcounter{section}{7}
\section{Appendix}
\label{sec:appendix}

\subsection{Experiments - Features}
For the experiments we restricted the lab tests to the following list of roughly 40 target harmonized features: blood pressure, pulse, respiratory rate, oxygen saturation, blood pressure mean, temperature, urine output, glucose, oxygen source, hematocrit, hemoglobin, potassium, creatinine, bun, sodium, chloride, platelet, wbc count, mcv, magnesium, calcium, phos, inr, pt, ptt, carbon dioxide, tbili, ast, alt, alkphos, lactate, ionized calcium, albumin, troponin, egfr mdrd, tsh, dbili, hgba1c, total cholesterol, HDL, LDL.
For the RNN model we used any codes that would map into one of these harmonized features.

\subsection{Experiments - Methods}
\paragraph{Dynamical Attention}
Attention-mechanisms provide a way to make recurrent neural networks more interpretable~\cite{Bahdanau2014-iu}. 
Attention derives weights for each intermediate state in the RNN by combining it with the final state. The prediction is then produced from the normalized weights sum across the states. 
Those weight can serve directly as explanations and have been used in the clinical context to identify important diagnosis codes~\cite{Sha:distillation,Retain-Attention}.

\paragraph{Odds Ratio and Rothman index}
Two statistics commonly used to associate risk with discretizied features, are the odds-ratio, e.g. in~\cite{Caruana:2015:Pneumonia},  and the Rothman index~\cite{Rothman2013-zx}. 

In particular, we compute the odds of the outcome for one of the ranges of a feature as the ratio of the number of values of the feature that fall into the range for examples with the outcome present vs. absent.
 We also compute the same odds considering values of this feature outside the particular range and take the ratio of the two odds.

The \emph{Rothman index} is defined as the ratio  of the empirical risk associated with a particular range of a feature over the empirical risk of the average feature value. 

To lift them to the dynamical setting, in addition to requiring the events to fall into the window of interest $(t_0, t_1]$, we also require those values to have changed since $t_0$. For example, when we try to explain a risk increase in the past 2 hours, it is not helpful to only point to events that happened more than 2 hours ago or values that have not changed since.
This second requirement is important in the case of stateless models generally, since the model does not have the ability to keep track of changes in the input. 

Rather than comparing the values of a feature directly which  is sensitive to variances across examples and within an example across time, we simply compare their associated weights. 
\begin{definition}
Given a sequence of attribution weights $a=a_0,\dots, a_{t_1}$, we define the \emph{time-diffed explanations} for the change from $p_{t_0}$ to  $p_{t_1}$  at time $t$ (with $t_0 < t \leq t_1$ ) as 
 \[ 
 a^{t_0\rightarrow t_1}_t = a_t - \max_{t' \leq t_0} a_{t'}.
 \]
 \end{definition}
Variants of this idea could replace the $\max$ over time with the weight of the most recent value until $t_0$. With time-diffing, explanations of a risk increase from $p_{t_0}$  to $p_{t_1}$ are focused on features whose odds ratio or Rothman index increased between $t_0$ and $t_1.$

\textbf{Properties:} 
Users can understand how the weights are being computed from simple counts over the dataset. 

The challenge of confounded irregular measurements is being addressed through normalization by considering the risk of the average lab result or the odds for other results from the same test.

\subsection{Experiments - Human Evaluation.}
The expert raters were asked to give their response on a 5-point Likert scale from ``Extremely Likely'' to ``Extremely Unlikely''. 
We consider explanations marked as ``Extremely likely'' or ``Very likely'' as correct.
The specific instructions accompanied by an example were as follows:

Your task is to to judge how much each test and value raise your concern that the patient's health has gotten worse since rounds	
On the right hand side are NEW values that have been collected since you last reviewed the patient's chart	
Specifically, for each value in column E-F (test and value), what is the likehood that this information in indicates that the patient's condition is worse since you last looked at the patients chart (i.e. during rounds).	
You should score your responses in column H from 1-5 with:	

\begin{enumerate}
    \item[5] Extremely likely (i.e. extremely likely to indicate a patient's condition is worse)
    \item[4] Very likely
    \item[3] Unlikely (does not help me assess more or less likely)
    \item[2] Very unlikely 
    \item[1] Extremely unlikely (i.e. this information likely indicates the patient's condition is better)
\end{enumerate}

\paragraph{Notes.}
The information in columns A-C is generally sorted by type (there are a few values out of order occasionally due to coding issues) to help you find the data easily
Please bias AWAY from picking 3  (explained below)
Choosing 5 does not necessarily mean you think the patient's risk has gone up a significant amount - only that the signal is clear that the risk has gone up unambiguously 
In the example above, the respiratory rate of 32 is scored at 4 because respiratory rates are known to be inconsistently recorded, so measurement error might account for that value
However, if a patient's lactate increased from 1.4 to 3 then you should score that as a 5 because it is clearer that the risk has actually increased.
Ignore all data not in columns A though H.